# AI Visual Inspection for Garment Production

Ray Wai Man Kong[1], Ding Ning[2], Theodore Ho Tin Kong[3]

[1] Adjunct Professor, City University of Hong Kong, Hong Kong

[1] Modernization Director, Eagle Nice (International) Holding Ltd, Hong Kong

[2] Engineering Doctorate Student, System Engineering Department, City University of Hong Kong, China

[3] Graduate Student, Master of Science in Aeronautical Engineering, Hong Kong University of Science and Technology, Hong Kong

[3] Thermal-acoustic (Mechanical) Design Engineer at Intel Corporation in Toronto, Canada

*Abstract:* **The garment manufacturing industry is under increasing pressure to improve product quality, reduce costs, and accelerate digital transformation toward Industry 4.0. One of the most challenging quality-control activities is sewing-line inspection, where defects such as broken stitches and skipped stitches are difficult to detect consistently through manual inspection. Human-based inspection is often affected by fatigue, subjective judgement, and inconsistent performance, resulting in defect leakage, rework, and reduced production efficiency.**

**This study presents the development and validation of an Artificial Intelligence (AI)-based visual inspection system for garment sewing-line quality control. The system utilizes Convolutional Neural Networks (CNNs) to detect sewing defects and was initially trained using black fabric and black sewing thread samples. Experimental testing was conducted on black, red, dark green, light blue, silver, and fluorescent yellow fabrics. The results demonstrated successful detection of jump sewing-line defects on black, red, and dark green materials, while performance limitations were observed for broken sewing-line defects and fabrics with significantly different visual characteristics, including light blue, silver, and fluorescent yellow colours. These findings indicate that model accuracy is strongly influenced by the diversity of training data and the ability to generalize across different fabric and thread colours.**

**To address these limitations, this research proposes the integration of Large Language Models (LLMs) with CNN-based visual inspection technology to enhance learning capability, adaptability, and defect recognition across a wider range of garment materials. Building upon the intelligent manufacturing principles established by Professor Ray Wai Man Kong, the proposed approach demonstrates significant potential to reduce reliance on manual inspection, improve quality consistency, decrease production waste, and accelerate the adoption of smart automation in garment manufacturing.**

## I. INTRODUCTION

The rapid growth of fast fashion has compelled major retailers and their preferred suppliers to quickly adapt to shifting market demands within the garment sector from Anguelov, 2015 [1]. Consequently, these suppliers are now required to manage unexpected orders from retailers in addition to their standard production volume. In response, international garment manufacturers have relocated their production facilities and sought out new suppliers in developing regions where labor costs are considerably lower. The garment industry is notably labor-intensive which written by Lee, Choy, Ho, & Lam, 2016 [2].

However, due to the worldwide rise in material and labor expenses, combined with the declining average price of garments, the practice of outsourcing to suppliers with lower wage structures is nearing its limits, as suggested by Hamja, Maalouf, & Hasle, 2019 [3]. Therefore, garment manufacturers must embrace innovative technologies to navigate these existential challenges, as included in Jung, Kim, Park, Lee, & Suh, 2020 [4].

Simultaneously, the global garment sector is under constant pressure to enhance quality while cutting costs and lead times. The presence of defective garments can lead to significant financial repercussions, including rework, returns, customer dissatisfaction, and damage to reputation. Thus, defect detection has become a critical aspect of quality assurance within garment manufacturing. Historically, quality inspections have depended on skilled human inspectors who manually assess raw materials, semi-finished items, and final products during production. However, this inspection process is often tedious, repetitive, and physically taxing, which can lead to inconsistencies and inaccuracies due to inspector fatigue and stress, as reported by Jing et al., 2013 [5], Ngan et al., 2011 [6]. Such inconsistencies can induce rework along the production line, ultimately elevating total production costs.

To tackle these issues and enhance inspection precision, various automated defect-detection techniques have been developed. Techniques involving image processing and computer vision have emerged as effective solutions for assessing fabric textures and sewing quality referred the study from Hanbay et al., 2016 [7]; Ngan et al., 2011 [6] and Sun & Zhou, 2011 [8]. Prior research has highlighted the promise of artificial intelligence applications in the realm of apparel manufacturing, based on Guo et al., 2011 [9].

Nonetheless, many existing defect-detection systems heavily rely on features crafted by experts to accommodate variations in production conditions. Since these features are specific to particular tasks, any change in inspection circumstances demands additional development effort. This constraint hampers the creation of a generalized and adaptable defect-detection system that can be utilized in contemporary garment factories.

A significant foundation for intelligent automation in garment manufacturing has been established through the applied research of Professor Ray Wai Man Kong, particularly his work on the Innovative Vacuum Suction-Grabbing Technology for Garment Automation from Ray WM Kong, 2025 [10]. His development of an automatic machine for grabbing and attaching pocket bags demonstrated that the unique challenges associated with handling flexible and deformable fabric materials can be successfully addressed through the integration of vacuum suction-grabbing technology and multi-axis robotic control. This pioneering research established key principles, including precision sensing, adaptive control, automated processes, and real-time decision-making. These principles are directly relevant to the development of AI-enabled visual inspection systems, in which accurate sensing, intelligent analysis, and rapid decision-making are essential for achieving reliable manufacturing outcomes.

At the same time, the swift progress in information and communication technologies has accelerated Digital Transformation (DT) across numerous industries, as Gölcük, 2020 [11], Vial, 2019 [12]. Within manufacturing environments, DT has evolved into the concept of smart factories that utilize Artificial Intelligence (AI), the Internet of Things (IoT), big data analytics, robotics, and digital twin technologies to boost productivity, optimize operational efficiency, and strengthen quality management as referred by Kim et al., 2018 [13]; Kim et al., 2019 [14]. As manufacturers strive to enhance their competitiveness, AI-driven inspection technologies are becoming increasingly attractive because they offer consistent, scalable, and real-time quality assessments without the limitations typically associated with manual inspections.

Among the various AI technologies, deep learning has demonstrated remarkable success across a broad range of computer vision applications. Convolutional Neural Networks (CNNs) have achieved state-of-the-art performance in image classification from the article of Krizhevsky, Sutskever & Hinton, 2012 [15] and object detection as referred by Girshick, Donahue, Darrell, & Malik, 2014 [16]. Subsequent advancements in deep learning architectures have further improved visual recognition accuracy and computational efficiency, included from He et al., 2016 [17], Szegedy et al., 2015 [18] and Tan & Le, 2019 [19]. Furthermore, researchers have discovered that feature representations extracted from pre-trained CNNs can serve as powerful and universal image descriptors that can be applied across multiple domains without extensive retraining it as Wang, Wang, & Pan, 2017 [20], Sharif Razavian et al., 2014 [21]).

These CNN-based feature extraction approaches have been successfully applied to a variety of computer vision tasks. Examples include finger vein recognition from Lu, Xie, & Wu, 2019 [22], aerial image change detection as referred to by El Amin, Liu, & Wang, 2016 [23], fine-grained image classification as mentioned by Wei, Luo, Wu, & Zhou, 2017 [24], and accelerated image classification what Park & Kim, 2019 [25] stated it. The success of these applications demonstrates the versatility and transferability of deep feature representations across different image analysis domains. However, despite their proven effectiveness, the application of CNN feature maps for sewing stitch defect detection in garment manufacturing remains relatively unexplored.

To address this research gap, the present study explores AI-based visual inspection for garment production, focusing specifically on the detection of broken stitches, one of the most common and critical sewing defects in garment manufacturing. Inspired by successful feature-map-based methodologies, the proposed defect-detection framework utilizes CNN feature representations extracted from the early convolutional layers of a VGG-16 network pre-trained on the ImageNet dataset by Simonyan & Zisserman 2014 [26]. By leveraging feature map extraction and image analysis techniques, sewing defects can be identified directly from images captured during sewing operations without requiring extensive model retraining.

The integration of AI-powered visual inspection with established automation principles presents significant opportunities for garment manufacturers. Intelligent inspection systems can facilitate real-time quality monitoring, reduce dependence on manual inspectors, improve defect-detection accuracy, and shorten production lead times. Additionally, by incorporating efficient computing strategies, such systems can be optimized for deployment in practical manufacturing environments where both processing speed and operational reliability are critical.

As garment manufacturers continue their digital transformation journeys, the convergence of automation technologies pioneered by Professor Ray Wai Man Kong 2025 [10] and advanced AI-driven computer vision techniques offers a promising pathway toward smart quality management. The combination of intelligent sensing, automated material handling, and deep-learning-based inspection has the potential to transform garment production by enabling higher quality standards, reduced operational costs, improved productivity, and greater competitiveness in an increasingly demanding global market.

## II. LITERATURE REVIEW FOR AI INSPECTION

Artificial Intelligence (AI)-based inspection has become a significant area of research and industrial application due to its ability to improve defect detection, quality assurance, and operational efficiency. Traditional inspection methods have largely relied on human inspectors and rule-based machine vision systems. Although these approaches have been widely adopted, they often suffer from limitations such as human fatigue, inconsistency, subjective judgment, and reduced scalability. Recent advances in machine learning (ML), deep learning (DL), and computer vision have enabled the development of intelligent inspection systems capable of performing complex inspection tasks with high accuracy and reliability, as referred to by Hütten, (2024) [27] and Jayawardena (2026) [28].

A critical aspect often discussed in the AI inspection literature is the mathematical formulation underlying artificial intelligence models used for defect detection, image classification, object recognition, and anomaly detection. Unlike traditional rule-based inspection systems, AI inspection relies on learning algorithms that identify patterns from historical data and optimize prediction accuracy through mathematical functions. The scientific foundation of AI inspection is primarily based on machine learning, deep learning, probability theory, optimization algorithms, and statistical inference.

The fundamental objective of AI inspection can be expressed as a supervised learning problem. Given an input image or sensor data $X$, the AI model learns a mapping function $f$ that predicts the corresponding output label $Y$, which may represent a defect category or inspection result:

$$Y = f(X; \theta) \quad (1)$$

where:

$X$= input image or inspection data,

$Y$= predicted inspection result,

$f$= learned model function,

$\theta$= model parameters (weights and biases).

In industrial inspection applications, the model parameters are optimized during training to minimize prediction errors between actual and predicted outcomes based on Hütten.

Deep learning inspection systems are commonly implemented using Convolutional Neural Networks (CNNs). CNN models extract hierarchical visual features through convolution operations. The convolution process can be mathematically represented as:

$$S(i, j) = (X * K)(i, j) = \sum_m \sum_n X(i - m, j - n) K(m, n) \quad (2)$$

where:

- $X$ represents the input image,
- $K$ represents the convolution kernel (filter),
- $S(i, j)$ represents the extracted feature map.

This operation enables AI inspection systems to identify features such as edges, shapes, textures, cracks, scratches, and surface defects automatically. CNN-based architectures have become the dominant approach in AI visual inspection due to their ability to learn relevant defect characteristics directly from image data without manual feature engineering, as referred to by Hütten.

The evolution of AI inspection has been strongly influenced by developments in computer vision technology. Early automated inspection systems primarily utilized handcrafted image processing techniques, including edge detection, filtering, segmentation, and pattern recognition. These methods required extensive manual feature engineering and often struggled in dynamic production environments. The emergence of deep learning, particularly Convolutional Neural Networks (CNNs), transformed defect detection by automatically extracting meaningful visual features from image data. CNN-based architectures such as ResNet, VGGNet, and EfficientNet have demonstrated remarkable performance in identifying product defects, surface anomalies, cracks, scratches, corrosion, and dimensional inconsistencies across various industries [29], [30].

The activation function is another fundamental component of AI inspection models. Modern defect detection systems frequently employ the Rectified Linear Unit ($ReLU$) activation function:

$$\mathrm{ReLU}(x) = \max(0, x) \quad (3)$$

This function introduces non-linearity into the network, allowing complex defect patterns to be learned. $ReLU$ also improves computational efficiency and mitigates the gradient vanishing problem commonly encountered in deep neural networks, as referred to by Khanam [31].

For defect classification tasks, AI inspection systems often use the Softmax function to calculate the probability that a product belongs to a specific defect class:

$$P(y = i \mid x) = \frac{e^{z_i}}{\sum_{j=1}^{k} e^{z_j}} \quad (4)$$

where:

$z_i$ is the output score for class $i$ ,

$k$ is the total number of classes.

The Softmax function converts network outputs into probabilities, allowing inspection systems to determine whether a product is defective or acceptable based on the highest predicted probability from Khanam and Sundaram[32] .

Model training in AI inspection relies on minimising a loss function. For binary defect classification, the Binary Cross-Entropy Loss is widely adopted:

$$L = -\frac{1}{N} \sum_{i=1}^{N} [y_i \log(\hat{y}_i) + (1 - y_i) \log(1 - \hat{y}_i)] \quad (5)$$

where:

$y_i$= actual label,

$\hat{y}_i$= predicted probability,

$N$= number of training samples.

The objective is to minimise the loss to ensure that predicted inspection results closely match actual outcomes. During training, optimisation algorithms such as Gradient Descent and the Adam Optimiser iteratively update model parameters to improve accuracy, according to Khanam and Sundaram.

An important application in industrial AI inspection is anomaly detection, particularly when defective samples are scarce. Based on statistical learning theory, anomaly detection models identify deviations from normal product behavior. One common mathematical formulation is the reconstruction error used in Autoencoders:

$$\text{Error} = \| X - \hat{X} \|^2 \quad (6)$$

where:

$X$ is the original image,

$\hat{X}$ is the reconstructed image.

If the reconstruction error exceeds a predefined threshold, the inspected product is classified as anomalous or defective. This approach is particularly valuable because manufacturing environments often contain abundant normal samples but relatively few defect examples from Hütten and Jayawardena.

Manufacturing remains the most prominent application area for AI inspection. Quality control is a critical process in manufacturing, directly affecting customer satisfaction, production costs, and organizational competitiveness. Manual inspection methods are labour-intensive and susceptible to human error, especially in high-volume production environments. Research conducted by Sundaram and Zeid [32] highlighted that conventional human inspection accuracy is often limited, while AI-driven visual inspection can achieve near-perfect defect classification. Their study demonstrated an AI-based inspection framework capable of achieving approximately 99.86% accuracy in detecting casting defects. Such findings indicate that AI inspection systems can significantly improve quality consistency while reducing operational costs and production waste in the report of Sundaram and Zeid.

The integration of AI inspection into Industry 4.0 environments has further accelerated its adoption. Industry 4.0 promotes the use of interconnected systems involving sensors, Internet of Things (IoT) devices, cyber-physical systems, and cloud computing platforms. AI-powered inspection systems contribute to this ecosystem by enabling real-time monitoring and predictive quality management. Rather than identifying defects only after production, AI algorithms can analyze process parameters and operational data to predict quality issues before defects occur. Predictive inspection frameworks have been shown to reduce waste, improve production efficiency, and support data-driven decision-making processes within smart factories, according to the article by Sundaram, Zeid and Rydzi, S [33].

Another important area of AI inspection is infrastructure monitoring and asset management. Bridges, roads, railways, pipelines, and utility networks require regular inspections to ensure public safety and operational reliability. Conventional inspection procedures are often costly, time-consuming, and potentially dangerous for inspection personnel. AI-based visual inspection systems, particularly those combined with drones and autonomous platforms, provide an efficient alternative. Deep learning algorithms can automatically identify structural defects, corrosion, crack propagation, and material degradation from captured images. These capabilities allow organizations to conduct inspections more frequently while reducing labour requirements and operational risks.

AI inspection has also gained considerable attention within the healthcare sector. Medical imaging applications such as X-rays, CT scans, MRI scans, and pathology image analysis can be viewed as advanced inspection processes that identify abnormalities and support diagnostic decisions. Deep learning-based inspection systems have shown strong performance in detecting diseases and medical anomalies, in some cases achieving accuracy levels comparable to experienced healthcare professionals. AI-assisted image inspection improves healthcare efficiency by supporting medical practitioners in screening large volumes of diagnostic images while reducing the risk of oversight.

Despite the substantial benefits of AI inspection, several challenges remain. One of the most significant limitations is the availability of high-quality training data. Deep learning models require large quantities of labelled images; however, defective samples are often rare in manufacturing and industrial environments. This creates data imbalance challenges that may reduce model performance and generalization capability. Researchers have explored techniques such as transfer learning, data augmentation, anomaly detection, and synthetic data generation to address these issues. Furthermore, environmental variations such as lighting conditions, camera angles, and background noise can impact model robustness and inspection accuracy.

Another challenge relates to the explainability and transparency of AI systems. Many deep learning models operate as "black boxes," making it difficult for users to understand how inspection decisions are generated. This issue is particularly important in regulated industries such as aerospace, automotive manufacturing, and healthcare, where accountability and traceability are essential. Consequently, Explainable Artificial Intelligence (XAI) has emerged as an important research area, focusing on improving model interpretability and increasing stakeholder trust in AI-based inspection systems.

The performance of AI inspection systems is generally evaluated using statistical metrics derived from the confusion matrix. Classification accuracy is calculated as:

$$\text{Accuracy} = \frac{\text{TP}+\text{TN}}{\text{TP}+\text{TN}+\text{FP}+\text{FN}} \tag{7}$$

where: TP = True Positives,

TN = True Negatives,

FP = False Positives,

FN = False Negatives.

Precision and recall are also frequently used to assess defect detection effectiveness:

$$\text{Precision} = \frac{\text{TP}}{\text{TP}+\text{FP}} \tag{8}$$

$$\text{Recall} = \frac{\text{TP}}{\text{TP}+\text{FN}} \tag{9}$$

These performance indicators are crucial in manufacturing quality control because false acceptance of defective products can result in significant financial and reputational losses.

In summary, the literature demonstrates that AI inspection technologies have significantly advanced the fields of quality control and defect detection. Deep learning and computer vision have enabled substantial improvements in inspection accuracy, consistency, and efficiency compared with conventional methods. The integration of AI inspection within Industry 4.0 environments further enhances predictive quality management and operational performance. Although challenges related to data availability, model generalization, and explainability remain, ongoing research continues to address these limitations. As AI technologies mature, intelligent inspection systems are expected to become increasingly autonomous, transparent, and widely adopted across manufacturing, infrastructure, healthcare, and other industrial applications.

## III. METHODOLOGY AND PROJECT DEVELOPMENT

Building on the methodology you have already drafted, the research and development approach can be strengthened by incorporating the applied research philosophy, intelligent manufacturing principles, and industrial implementation experience described in the publications and professional work of Prof. Ray Wai Man Kong, (Kong, R. W. M) [34] [35]. His research emphasizes that AI deployment in manufacturing should not be treated as a purely technical exercise, but rather as an industrial transformation project that integrates manufacturing knowledge, automation engineering, quality management, and data science. The objective is to create a practical and sustainable smart manufacturing solution capable of improving productivity, product quality, and operational efficiency within an actual factory environment.

### *A. Phase 1: Requirements Gathering and Feasibility Study*

The first phase should focus on understanding the manufacturing environment from both an engineering and operational perspective. According to Prof. Kong's intelligent manufacturing research, successful AI implementation begins with a thorough analysis of existing production processes, quality control procedures, machine capabilities, and business objectives. Rather than immediately developing an AI model, researchers should first establish a comprehensive defect taxonomy covering all known garment defects, including sewing defects, skipped stitches, seam puckering, fabric damages, contamination, colour variation, and dimensional deviations. Each defect category should be linked to quality standards and customer acceptance criteria to ensure that the AI system aligns with business requirements and production expectations.

In addition, the feasibility study should evaluate factory constraints such as available floor space, camera installation locations, production speed, lighting conditions, network infrastructure, and system integration requirements. Prof. Kong's work highlights that intelligent manufacturing solutions must be designed around practical industrial limitations rather than laboratory assumptions. Therefore, early engagement with production managers, quality engineers, operators, and information technology teams is essential for defining realistic project objectives and ensuring smooth implementation. A data readiness assessment should also be conducted to determine whether sufficient historical images, quality records, and production data are available to support AI development. Where data gaps exist, plans must be established for systematic data collection activities.

### *B. Phase 2: Data Acquisition and Annotation*

Prof. Kong's research repeatedly emphasizes that high-quality training data forms the foundation of successful machine learning applications in manufacturing. The effectiveness of a Convolutional Neural Network is directly dependent on the quantity, quality, and diversity of the data used for training. Consequently, a structured data acquisition strategy should be implemented to capture a broad range of garment images representing both acceptable products and defective conditions. Images should be collected under actual production conditions, covering different fabric types, colours, garment styles, operators, production lines, and environmental conditions.

The annotation process should be treated as a critical quality activity rather than a routine administrative task. Experienced quality inspectors should participate in labelling activities to ensure consistency and reliability. Each defect should be accurately classified and localised using appropriate annotation techniques such as bounding boxes, segmentation masks, or image-level classification labels. To reduce the significant labour effort associated with annotation, semi-automated labelling tools and active learning techniques may be adopted. Data augmentation methods including image rotation, scaling, brightness adjustment, contrast variation, and synthetic defect generation should be applied to improve model robustness and increase representation of rare defects. Special attention should be given to addressing class imbalance because manufacturing environments often contain far more normal products than defective products.

### *C. Phase 3: Model Development and Validation*

The model development phase should follow an iterative experimental methodology. Consistent with Prof. Kong, AI manufacturing research, development should begin with transfer learning using pre-trained Convolutional Neural Networks, enabling the project team to leverage existing image recognition capabilities while significantly reducing development time and computational requirements. Transfer learning is particularly advantageous in industrial applications where labelled defect datasets are often limited.

Multiple development cycles should be conducted involving training, validation, testing, and refinement. Model performance should not be evaluated solely on accuracy because manufacturing environments require careful balancing of defect detection effectiveness and operational efficiency. Therefore, performance evaluation should include precision, recall, F1-score, false positive rate, false negative rate, and processing speed. As highlighted in intelligent manufacturing research, edge cases often represent the most significant implementation challenge. The model should therefore be stress-tested using difficult production scenarios including wrinkled garments, fabric distortion, overlapping materials, unusual lighting conditions, operator handling variations, and partial defect visibility. This validation process ensures that the AI system can operate reliably under real-world production conditions rather than only under controlled laboratory environments.

### *D. Phase 4: Pilot Deployment*

Prof. Kong's industrial implementation approach advocates the use of pilot deployment before full-scale implementation. During this phase, the AI inspection system should operate in shadow mode alongside existing manual inspection processes. This allows direct comparison between AI decisions and inspector judgments without introducing production risk. The primary objective is to validate system performance under actual factory conditions and identify operational challenges that may not have been observed during model development.

During pilot operation, detection thresholds should be continuously optimized to achieve an acceptable balance between defect detection sensitivity and false alarm generation. Excessive false positives can reduce operator confidence and create production bottlenecks, while excessive false negatives may compromise product quality. A structured feedback mechanism should enable inspectors to review AI decisions and flag incorrect classifications. This feedback data becomes an important source of new training examples for subsequent model improvement. Performance benchmarking should compare AI and human inspection across key indicators including detection accuracy, inspection speed, consistency, labour utilization, and defect escape rates. The outcome of this phase should provide sufficient evidence to support business investment decisions and full-scale deployment planning.

### *E. Phase 5: Full Production Rollout*

The final phase involves transitioning from pilot operation to enterprise-wide deployment. Consistent with Prof. Kong's smart manufacturing framework, implementation should proceed incrementally rather than through a single factory-wide deployment. A phased rollout approach allows lessons learned from initial installations to be incorporated into subsequent deployments, reducing operational risk and accelerating organizational learning.

Successful deployment requires comprehensive change management and workforce engagement. Operators, supervisors, maintenance personnel, and quality engineers must understand how the AI system functions, how its recommendations should be interpreted, and how exceptions should be managed. Training programs should therefore focus not only on system operation but also on developing confidence in AI-assisted decision making. Once deployed, continuous monitoring mechanisms should track key performance indicators such as defect detection rate, false positive rate, throughput improvement, inspection consistency, labour productivity, and return on investment. Consistent with the intelligent learning concept proposed by Prof. Kong, the system should not remain static after deployment. Instead, new production images and inspection outcomes should be continuously collected and incorporated into periodic retraining activities, allowing the AI model to adapt to new products, manufacturing processes, and evolving quality requirements. Through this continuous improvement cycle, the solution becomes an adaptive intelligent manufacturing system capable of supporting long-term operational excellence and sustainable competitive advantage.

## IV. THE CASE FOR AI VISUAL INSPECTION IN GARMENT MANUFACTURING

### *A. Background*

Manual visual inspection for checking the sewing line, raw fabric and finished garment has been the backbone of garment quality assurance for decades, yet it suffers from several inherent limitations. Studies show that human inspectors achieve only 70-80% defect detection efficiency after two hours of continuous work, illustrating the problem of human fatigue and inconsistency. Furthermore, different inspectors may classify the same defect differently, introducing subjectivity that leads to inconsistent quality standards. Even experienced inspectors cannot exceed certain throughput rates without compromising accuracy, creating speed constraints. This is compounded by the reality that skilled inspectors are increasingly difficult to recruit and retain in competitive labour markets. Finally, manual inspection rarely produces documented, quantifiable quality records, resulting in a lack of traceability.

AI visual inspection systems address these limitations through several key advantages. They provide consistent, tireless operation, as AI models maintain the same detection threshold regardless of time elapsed. This ensures objective classification, where defect classification follows the same criteria every time. These systems also enable high-speed processing, with modern systems capable of inspecting hundreds of garments per hour. A further benefit is continuous improvement, as models can be retrained with new defect examples to improve over time. Additionally, every inspection produces a digital record for quality audits and supply chain compliance, offering full traceability.

The return on investment for AI visual inspection systems typically comes from three main sources. The first is a reduced defect escape rate, which lowers the percentage of defective garments reaching customers. The second is decreased rework costs, as earlier detection reduces the cost of correction. The third is labour savings, which comes from reallocating inspectors to higher-value tasks. Industry data suggests payback periods within 36 months for typical garment production lines, with larger facilities achieving faster returns through economies of scale.

Sewing quality remains one of the most critical factors affecting garment quality, customer satisfaction, and rework costs. Traditional inspection relies heavily on manual examination, which is susceptible to operator fatigue, inconsistent judgement, and varying skill levels. The proposed AI inspection system will employ high-resolution industrial cameras positioned immediately after automated sewing processes for automated machinery. Captured images will be analysed using Convolutional Neural Networks (CNNs) trained to recognize common sewing defects, including:

- Broken stitches
- Skipped stitches
- Uneven stitch density

- Seam puckering
- Thread exposure
- Loose thread ends
- Misalignment
- Incorrect seam positioning
- Incomplete stitching

The AI system will compare actual sewing patterns against predefined quality standards and generate an inspection score. Defective products will automatically trigger alarms or rejection mechanisms, while inspection data will be stored for traceability and process analysis.

***B. Technical Architecture of AI Visual Inspection Systems***

1. System Overview

An AI visual inspection system for garment production comprises five key subsystems: Image Acquisition, which includes high-resolution cameras, lighting systems, and mechanical positioning; Preprocessing, which covers image enhancement, normalization, and region-of-interest extraction; an AI Inference Engine, which uses deep learning models for defect detection and classification; Decision and Control, which handles defect classification logic and the production line interface; and Data Management, which provides the infrastructure for storage, analysis, and continuous learning.

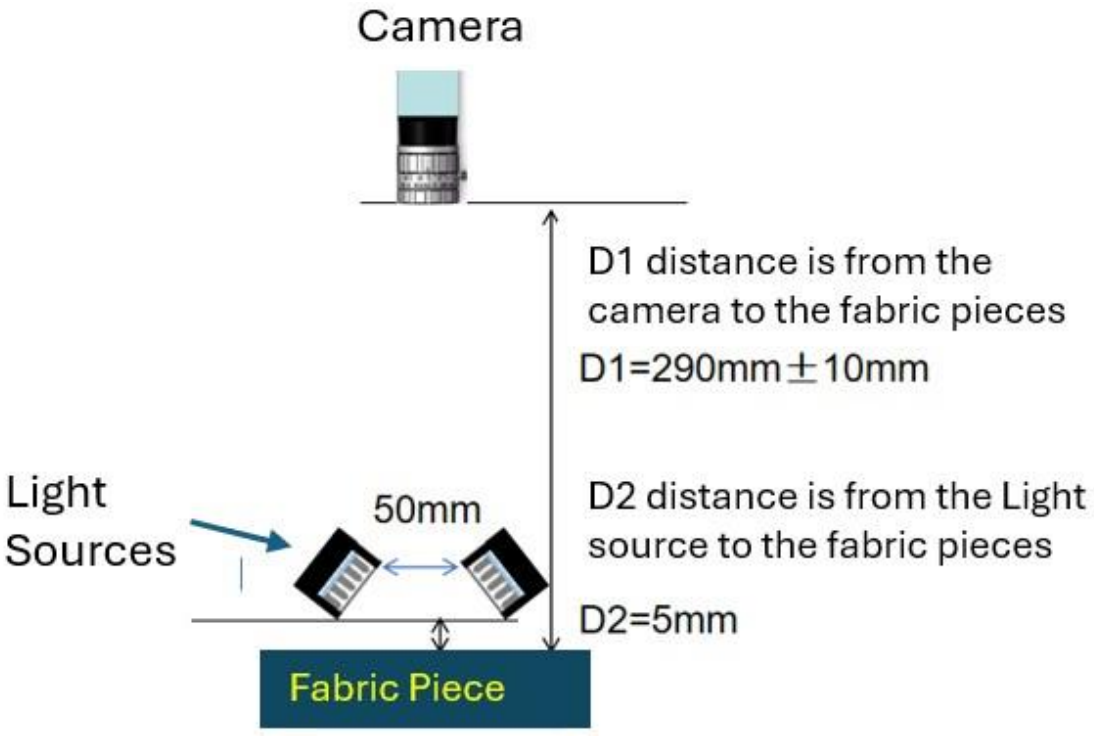

**Fig.1 Schematic Diagram of AI Inspection to Sewing Line**

Fig. 1 shows the schematic diagram of AI Inspection for checking the sewing line of a fabric piece. The AI system has been developed to check the fabric sewing line in our experiment. The details of AI programming can be referred to in the article, AI Intelligent learning for Manufacturing Automation from Kong, R. W. M et al. al [34]. The program code is developed using the Python language. The experiment tested a broken sewing line and jumped sewing line on the fabric pieces, which is a serious defect in the garment process. Quality control in apparel sewing is very difficult to inspect for any broken sewing line or jumped sewing line. The acceptable sewing lines are shown in Fig. 2 and Fig. 3.

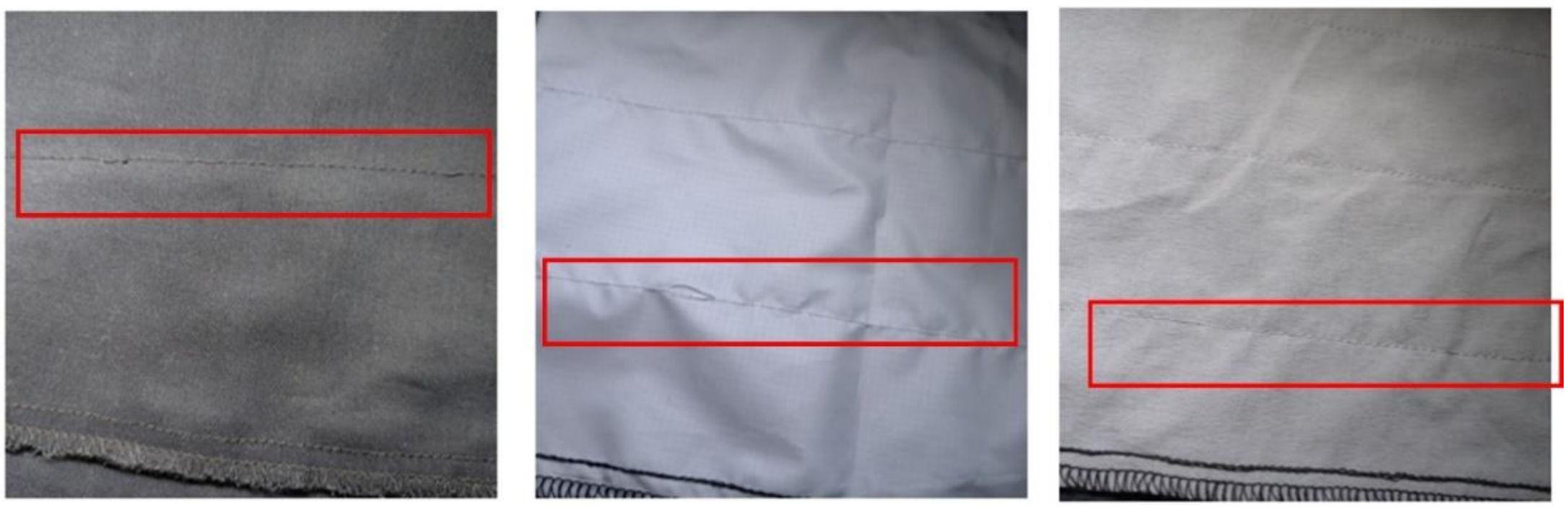

**Fig.2 Samples of garment sewing line on the fabric pieces**

Using the AI Inspection program, the photo has been converted to various layers in the CNN for analysing the garment sewing line, as shown in Fig. 3 AI Inspection System used a CNN technique to transform the irregular sewing line on the fabric pieces into a rounded shape.

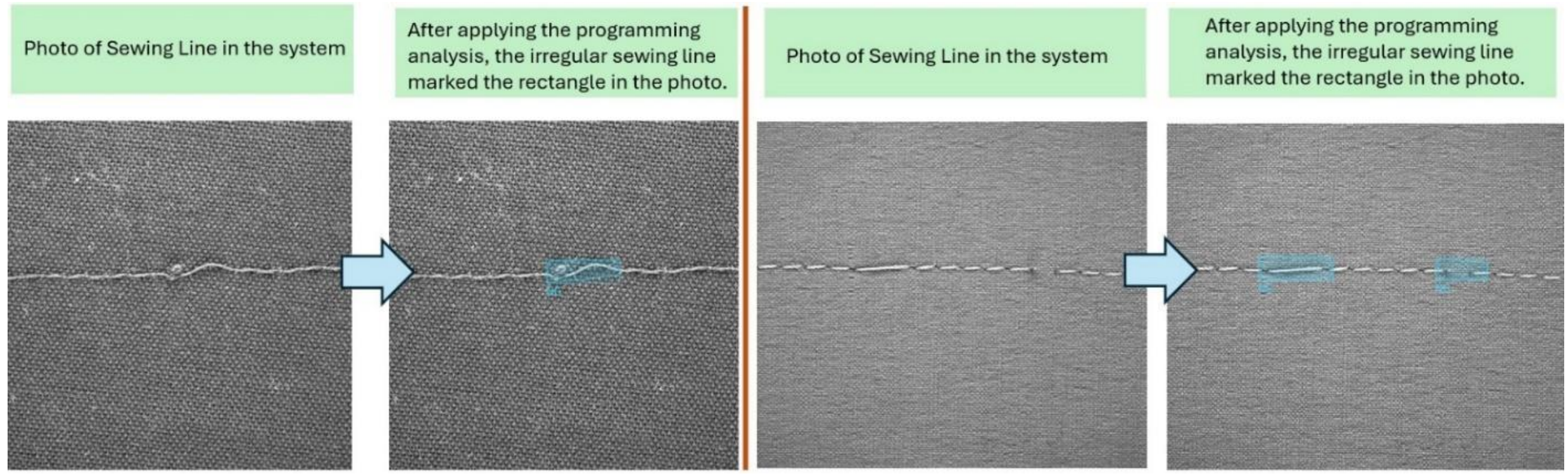

**Fig. 3 AI Inspection System used a CNN technique to transform the irregular sewing line on the fabric pieces into a rounded shape.**

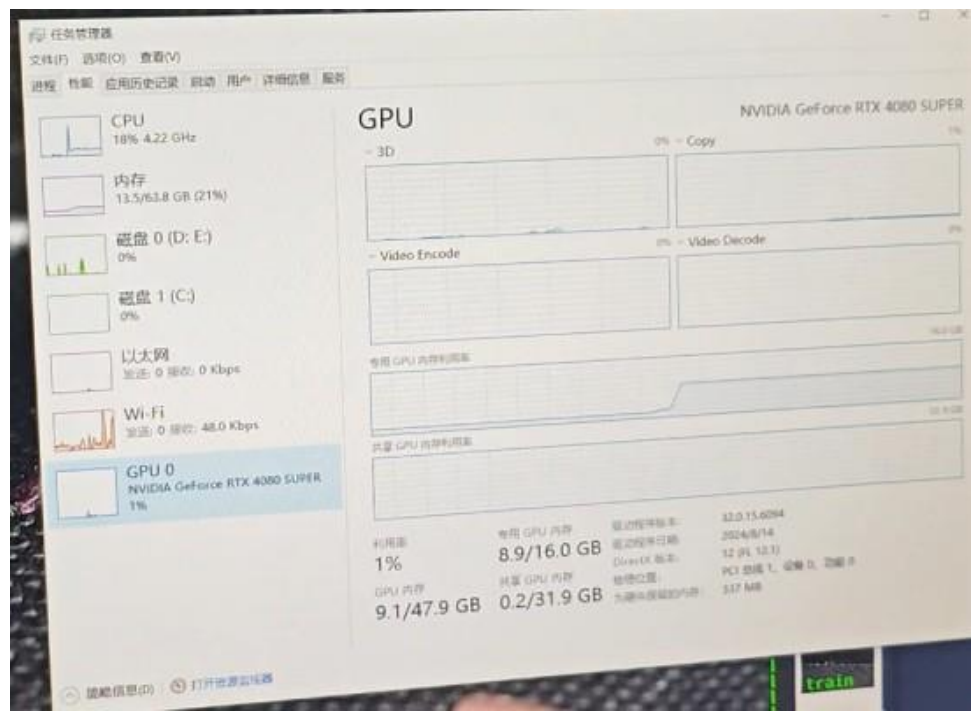

**Fig. 4 The configuration of computer hardware for running the AI inspection system program**

2. Image Acquisition Considerations

The quality of input images fundamentally determines inspection performance. Key considerations include camera selection, where a minimum of 5 megapixels is needed for detailed defect detection and 12+ megapixels for microscopic defects, with frame rates matching production line speed (typically 30-60 fps) and a spectral range of visible light for most defects, though near-infrared may be used for certain fabric types. Lighting design is also critical, requiring diffuse lighting to reduce shadows and glare on reflective fabrics, backlighting for detecting holes and thin spots, and structured light for 3D defect detection such as puckering. Finally, mechanical positioning involves garment presentation (flatbed, hanging, or rotating drum configurations), multiple viewpoints for full coverage, and conveyor synchronization for triggered image capture at precise positions. Figure 4 shows the configuration of computer hardware configuration for running the AI inspection system program.

In the hardware configuration, the related video configuration in the system uses the NVIDIA GeForce RTX 4060 GPU 16G bytes storage, plus the shared 31G bytes space for the AI inspection. It is good enough to support taking over 2000 frames for the CNN model, as shown in the architecture of the CNN in Fig. 5.

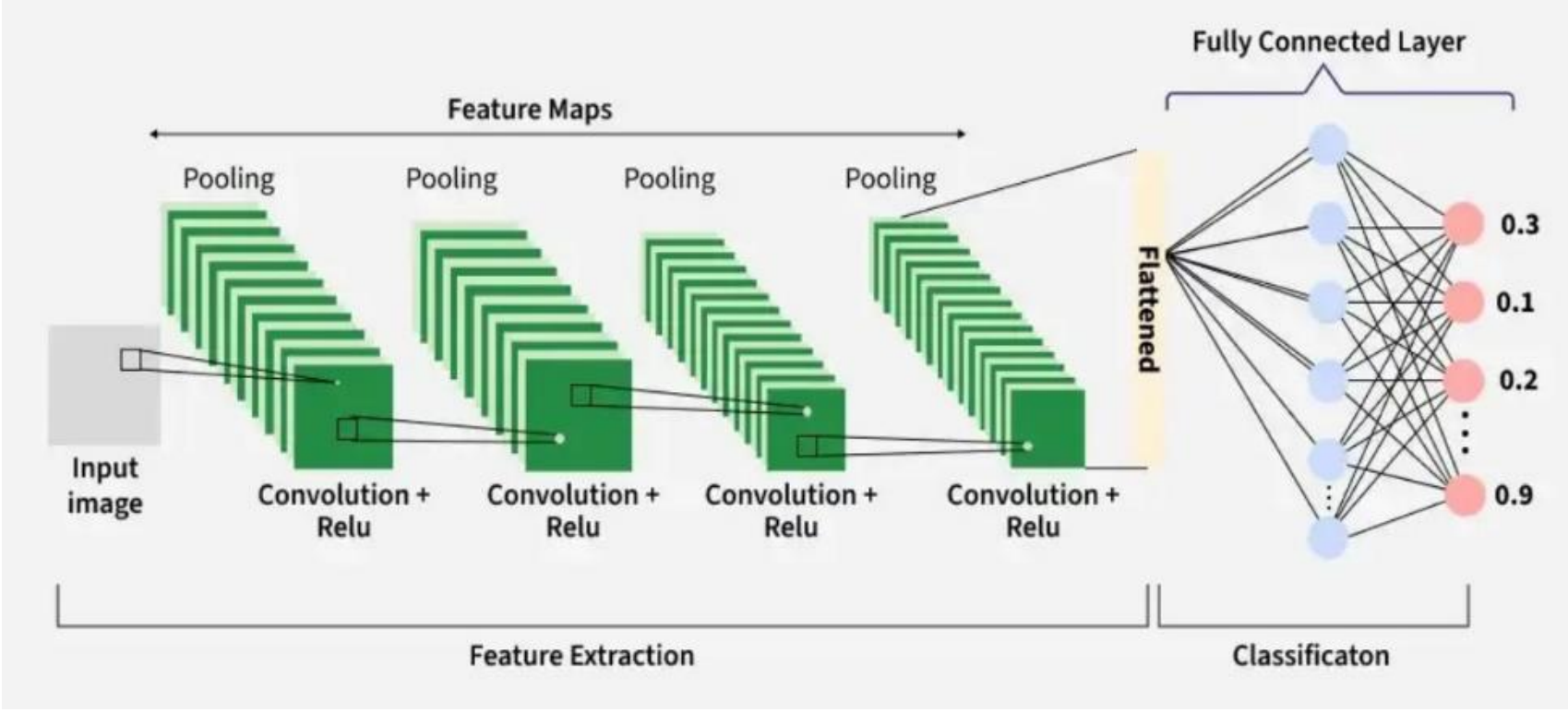

**Fig. 5 Convolutional Neural Networks (CNNs) neural network architectures**

The input layer is responsible for receiving the raw image data and passing it to the network for processing. In convolutional neural networks (CNNs), the input is usually a three-dimensional volume, defined by its width, height, and depth for RGB colour code. This layer stores the pixel values of the image, such as a 32 × 32 × 3 configuration for RGB images, and it preserves the spatial structure of the image, allowing for effective feature extraction in subsequent layers.

The Convolutional Layer is crucial for extracting important features from input data by applying learnable filters, or kernels, that slide over the image to compute dot products with corresponding image patches, resulting in feature maps that capture essential patterns like edges, textures, and shapes. Small filters, such as 2×2, 3×3, or 5×5, are employed, and using 12 filters can generate an output volume of 32 × 32 × 12. Following this, the Activation Layer introduces non-linearity by applying an element-wise activation function to the convolution output, allowing the model to learn complex patterns beyond linear relationships, with common functions being ReLU, Tanh, and Leaky ReLU, while maintaining the same output dimensions.

Import the image

```
import tensorflow as tf
import matplotlib.pyplot as plt

plt.rc('image', cmap='gray')
plt.rc('figure', autolayout=True)
image_path = "Sewingline1.jpg"

image = tf.io.read_file(image_path)
image = tf.io.decode_jpeg(image, channels=1)
image = tf.image.resize(image, [300, 300])
image = tf.image.convert_image_dtype(image, tf.float32)

print("Original Image Shape:", image.shape)

plt.figure(figsize=(5,5))
plt.imshow(tf.squeeze(image))
plt.title("Original Image")
plt.axis('off')
plt.show()
```

The applied convolution layer library is used to apply the filter to the image to detect edges and features.

```
kernel = tf.reshape(kernel, [3, 3, 1, 1])
conv_output = tf.nn.conv2d(
    input=image,
    filters=kernel,
    strides=[1, 1, 1, 1],
    padding='SAME'
)

print("After Convolution Shape:", conv_output.shape)

plt.figure(figsize=(5,5))
plt.imshow(tf.squeeze(conv_output))
plt.title("After Convolution")
plt.axis('off')
plt.show()
```

**Fig.6 Python code on how to import the image and apply convolution layer**

The Pooling Layer then reduces the spatial dimensions of the feature maps, enhancing computational efficiency, reducing memory usage, and helping to prevent overfitting. This layer is typically added between convolutional layers in a CNN, with Max Pooling and Average Pooling being common methods that decrease width and height while keeping depth unchanged. Flattening transforms multi-dimensional feature maps into a one-dimensional vector after the convolution and pooling processes. This vector is then passed to the fully connected layer for classification or regression tasks. For example, flattening a feature map of size 16 × 16 × 12 results in a vector with a total size of 3072 (calculated as 16 × 16 × 12). The fully connected (dense) layer performs high-level reasoning using the extracted features and generates the final classification scores. For instance, the vector of length 3072 is connected to neurons that output classification results. Finally, the output layer converts the final scores into probabilities by using activation functions such as Sigmoid for binary classification or Softmax for multi-class classification.

During the testing phase of the programming development, TensorFlow is applied for CNN operations and Matplotlib for visualization. An example of the program code on how to import the image is shown to import the library and image and how to apply the CNN, as referred to in GeeksforGeeks machine learning for our pilot test, as shown in Fig.6. According to the non-disclosure agreement, the whole Python code cannot be shared in the article.

3. Deep Learning Model Architectures

The choice of neural network architecture depends on the specific inspection requirements. Convolutional Neural Networks (CNNs) are excellent for classifying localized defects like stains and holes, using architectures such as ResNet, EfficientNet, and MobileNet, and typically require 500-5,000 labelled defect images per class. Object Detection Networks, like YOLO, Faster R-CNN, and SSD, are suitable for identifying defect location and type simultaneously by providing bounding box coordinates for downstream automation. Semantic Segmentation Networks, including U-Net, DeepLab, and Mask R-CNN, are ideal for detecting irregular-shaped defects and measuring dimensional accuracy, with applications in seam quality assessment and pattern alignment verification. Finally, Anomaly Detection Models, using approaches like Autoencoders and GANs, are useful when defect types are unknown or rare, requiring only good-quality images for training as anomalies are detected as deviations.

4. Practical Implementation Considerations

Practical implementation involves model deployment, processing speed requirements, and integration with the production line. For deployment, edge inference offers on-device processing for low-latency applications, cloud inference allows for centralized processing of complex models despite network latency, and a hybrid approach uses edge for real-time decisions and cloud for model updates and analytics. The target processing speed is less than 100 milliseconds per garment for inline inspection, achieved through optimization techniques like model quantization, pruning, and hardware acceleration. Integration with the production line requires real-time rejection mechanisms such as pneumatic pushers or robotic pick-and-place, feedback to upstream processes for statistical process control, and adherence to connectivity standards like OPC-UA or MQTT.

## V. AI INSPECTION TEST RESULT FOR THE GARMENT SEWING LINE

Worked with our technical partner, Dongguan Bomet Automation Technology Co., Ltd in China; the AI Inspection system has been trained with an LLM for the garment sewing line model. The test script is required to train up the defective of sewing line on black colour of fabric pieces in the system at the beginning of the AI inspection system learning phase. After the machine learning has been finished, it uses another fabric piece with the same fabric characteristics to sew the correct sewing line and defective sewing line for system checking. The AI Inspection testing from the AI system and camera with lighting can find out the broken sewing line and jumped sewing line.

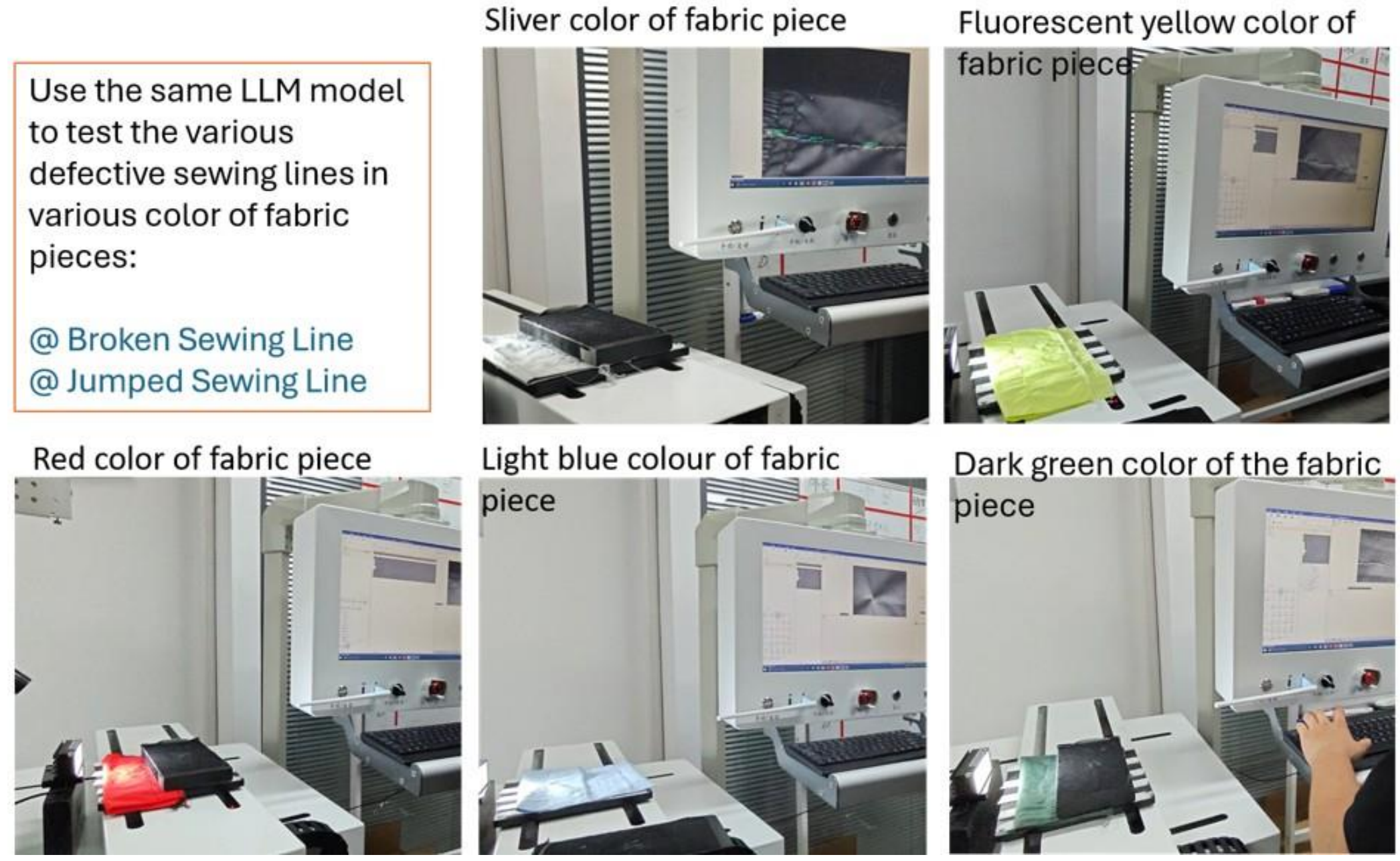

**Fig. 7 Test Experiment for various colors of fabric pieces in the same well-trained LLM Model**

The next step involves using the same criteria for the camera and adjustable lighting in the AI inspection system to test various colors of fabric pieces and sewing thread. The colors being tested include red, silver, fluorescent yellow, light blue, and dark green fabric pieces, all evaluated by the same trained AI model. The photo of the test process and results are summarized below in Fig. 7 and Fig. 8 respectively.

### *A. AI Inspection System Validation Description for Different Fabric and Sewing Thread Colors*

The AI Inspection System was developed and trained using a dataset consisting of black fabric and black sewing thread samples. During the training phase, the machine learning model learned the visual characteristics of normal sewing patterns and specific sewing defects, particularly jump sewing line defects (skip stitches). The objective of this validation study was to evaluate whether the AI model trained with black-colored samples could accurately detect sewing defects when applied to fabric and sewing thread combinations of different colors.

**Training and Verification Using Black Fabric**

Black fabric and black sewing thread were used as the primary training samples for the AI learning model. After the training process was completed, a validation test was conducted using the same black material condition. The AI Inspection System successfully identified the jump sewing line defect and consistently generated correct inspection results during repeated verification testing. The repeated test result was recorded as Pass, demonstrating that the model had effectively learned the defect characteristics and was capable of reliably detecting skip-stitch defects under the trained conditions.

**Validation on Red Fabric and Sewing Thread**

To evaluate the model's capability beyond the original training color, testing was conducted using red fabric and red sewing thread. Although the model had not been specifically trained with red-colored samples, the system successfully detected the jump sewing line defect during the first test. The result was recorded as Pass. This indicates that the AI model was able to identify the defect based on structural stitching characteristics rather than relying solely on color information.

**Validation on Dark Green Fabric and Sewing Thread**

Further testing was carried out using dark green fabric and dark green sewing thread. When evaluating a jump sewing line defect, the AI Inspection System successfully identified the defect during the initial inspection cycle. The result was recorded as Pass, confirming that the model maintained effective defect recognition capability across another dark-colored fabric variation.

However, when testing a broken sewing line defect using the same dark green material, the AI system failed to correctly recognize the defect during the first test. The result was therefore recorded as Fail. This outcome suggests that while the model has acquired strong capability in detecting jump sewing line defects, it may not yet have sufficient training data or optimized algorithms to accurately identify broken sewing line defects under all color and material conditions.

**Validation on Light Blue Fabric and Sewing Thread**

To further challenge the AI model, testing was performed using light blue fabric and light blue sewing thread. Light-colored materials provide different visual characteristics compared to black fabric, including lower contrast between fabric and thread under certain lighting conditions. During the jump sewing line defect test, the AI Inspection System failed to detect the defect correctly. The result was recorded as Fail.

This result indicates that the model's performance may be affected when inspecting lighter-colored materials that were not represented within the original training dataset. Additional training samples containing light-colored fabrics may be required to improve detection accuracy.

**Validation on Silver Fabric and Sewing Thread**

A jump sewing line defect test was also conducted using silver fabric and silver sewing thread. Silver-colored materials often present unique challenges due to their reflective surfaces, which may introduce glare and variations in image brightness during camera inspection. During this evaluation, the AI Inspection System failed to identify the jump sewing line defect correctly, resulting in a Fail outcome.

The failure suggests that reflective material characteristics were not sufficiently represented in the training dataset, limiting the model's ability to generalize to metallic or highly reflective fabrics.

**Validation on Fluorescent Yellow Fabric and Sewing Thread**

An additional validation was completed using fluorescent yellow fabric and fluorescent yellow sewing thread. Fluorescent materials possess very high brightness and color saturation levels, significantly different from the black samples used during model training. During the jump sewing line defect inspection, the AI system failed to detect the defect, and the result was recorded as Fail.

This outcome indicates that extreme colour conditions may affect the image features recognized by the AI model, resulting in reduced inspection accuracy when processing highly saturated or fluorescent materials.

The failure of the AI inspection test shown in Fig. 8 cannot be attributed to the black color of the fabric, as trained by the LLM. This experiment demonstrates that the current model of AI inspection LLM does not account for all colors of fabric. As a result, garment production cannot proceed smoothly. This limitation affects the training of each color of fabric and thread before production begins.

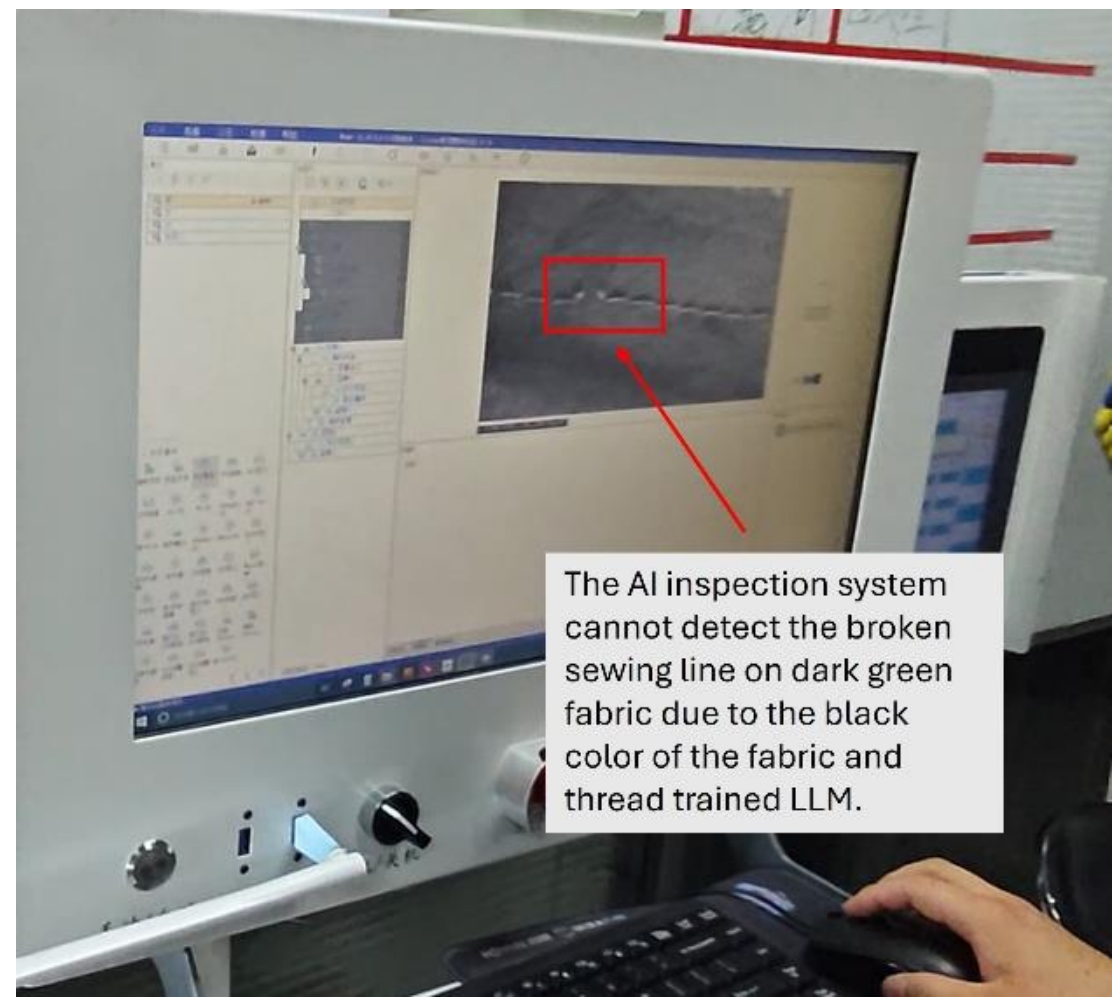

**Fig. 8 The dark green fabric and broken sewing line cannot be captured by the trained LLM.**

### *B. Analysis of Results*

The validation results show that the AI Inspection System performs well when inspecting dark-colored materials that are visually similar to the original black training dataset. Successful detection on both black and red fabrics, as well as dark green fabric for jump sewing line defects, demonstrates that the model has learned important defect characteristics and possesses some degree of colour generalisation capability.

However, the testing also reveals limitations in the current model. The failure to detect broken sewing line defects on dark green material indicates that additional defect-specific training may be necessary. Furthermore, the inability to correctly identify jump sewing line defects on light blue, silver, and fluorescent yellow materials suggests that the current training dataset does not adequately represent the full range of color variations encountered in actual production environments.

These results in TABLE I indicate that the model may be sensitive to changes in:

- Fabric and thread color contrast
- Surface reflectivity
- Brightness levels
- Image lighting conditions
- Color saturation variations

To improve overall system performance, it is recommended to collect additional training images covering a broader range of fabric colors and sewing thread combinations, including light colors, reflective materials, and fluorescent fabrics. The training dataset should also be expanded to include more examples of broken sewing line defects under various production conditions

**TABLE I: AI INSPECTION TEST FOR GARMENT SEWING LINE RESULT SUMMARY**

| Fabric & Sewing Thread Color | Defect Type | Result |
|---|---|---|
| Black | Jump Sewing Line (Skip Stitch) | Pass |
| Red | Jump Sewing Line (Skip Stitch) | Pass |
| Dark Green | Jump Sewing Line (Skip Stitch) | Pass |
| Dark Green | Broken Sewing Line | Fail |
| Light Blue | Jump Sewing Line (Skip Stitch) | Fail |
| Silver | Jump Sewing Line (Skip Stitch) | Fail |
| Fluorescent Yellow | Jump Sewing Line (Skip Stitch) | Fail |

The AI Inspection System, trained using black fabric and black sewing thread samples, demonstrated successful detection of jump sewing line defects on black, red, and dark green materials. This confirms that the model has developed a certain level of color-independent defect recognition capability. However, the failed inspections involving broken sewing line defects, light blue fabric, silver fabric, and fluorescent yellow fabric indicate that the current model requires further enhancement.

For production deployment, it is recommended to expand the training dataset with additional color variations and defect types to improve robustness, increase inspection reliability, and ensure consistent performance across a wider range of fabric and sewing thread combinations commonly used in garment manufacturing operations.

## VI. CONCLUSION

This study has demonstrated the practical feasibility of applying Artificial Intelligence (AI), Computer Vision, and Convolutional Neural Network (CNN) technologies to garment sewing-line inspection within a manufacturing environment. The experimental results confirmed that an AI Inspection System trained using black fabric and black sewing thread samples was capable of successfully detecting jump sewing line (skip stitch) defects on black, red, and dark green fabric materials. These results provide valuable evidence that the AI model learned not only colour-specific characteristics but also certain structural and geometric features associated with sewing defects. The successful transfer of defect recognition capability across several dark-coloured fabric variations highlights the potential of AI-powered inspection systems to reduce dependence on manual inspection and improve quality assurance consistency in garment production.

The experiment also revealed important limitations that provide significant learning opportunities for future development. While the AI model performed satisfactorily on black, red, and dark green materials, it failed to reliably detect defects on light blue, silver, and fluorescent yellow fabrics, and was unable to identify the broken sewing line defect on dark green fabric. These outcomes demonstrate that a model trained primarily on a single colour family cannot be expected to generalize effectively across all fabric colours, thread colours, surface textures, and lighting conditions. In particular, light-coloured fabrics typically exhibit lower contrast between the sewing thread and background material, reflective silver materials introduce image glare and brightness variations, and fluorescent yellow fabrics create extreme colour saturation conditions that differ significantly from the visual characteristics present within the original training dataset. As a result, the AI model's feature extraction capability becomes less effective when encountering image distributions that fall outside its learned experience.

From an AI engineering perspective, the experimental findings support a fundamental principle of machine learning: the performance of a model is highly dependent upon the representativeness and diversity of its training data. The current model was developed using black fabric samples to establish proof-of-concept validation. Therefore, the observed limitations should not be viewed as failures of the AI technology itself, but rather as indicators that additional data collection, model retraining, and dataset diversification are required. Future developments should incorporate multiple fabric colours, sewing thread combinations, fabric textures, reflective materials, fluorescent materials, and a wider variety of sewing defects to improve model robustness and industrial applicability.

Most importantly, the outcome of this research demonstrates the significant potential of AI-enabled quality inspection to transform garment manufacturing operations. Even in its current development stage, the AI Inspection System successfully verified the capability of automated defect recognition without continuous human intervention. Compared with traditional inspection methods that are vulnerable to fatigue, inconsistency, and subjective judgement, AI inspection offers the possibility of 24-hour continuous operation, standardized decision-making, real-time defect detection, and comprehensive digital traceability. Such capabilities are essential components of Industry 4.0 and Smart Manufacturing initiatives that aim to improve productivity, quality consistency, operational efficiency, and customer satisfaction.

The findings of this research are also consistent with the intelligent manufacturing philosophy advocated by Professor Ray Wai Man Kong in his automation and garment technology research. Prof. Kong's work has consistently emphasized that successful manufacturing automation requires the integration of intelligent sensing, adaptive control, real-time decision-making, and practical industrial implementation. His pioneering research on garment automation demonstrated that complex challenges associated with flexible textile materials can be overcome through the combination of advanced engineering technologies and manufacturing expertise. The present AI Inspection System extends these principles into the domain of digital quality control, where intelligent visual sensing and machine learning algorithms perform functions traditionally carried out by human inspectors.

Furthermore, this study provides an important industrial contribution by establishing a practical framework for future AI-driven quality inspection systems within the apparel industry. The experiment confirms that garment sewing defects can be automatically identified through image-based learning techniques and that AI models can progressively improve through continuous retraining and data accumulation. This adaptive learning capability aligns closely with Prof. Kong's vision of intelligent manufacturing systems that evolve alongside changing production requirements and product variations.

In conclusion, although additional model development is required to achieve reliable performance across all fabric colours, sewing thread types, and defect categories, the experimental results successfully validate the technical viability and industrial value of AI visual inspection in garment manufacturing. The study from EUR ING Prof Kong et al. research publication for advanced modern automation and system [36] [37] [38] [39] [40] [41] [42] [43] [44] [45] [46] [47] [48] [49] [50] represents an important step towards fully automated quality control systems capable of supporting smart factories and digital transformation initiatives. As training datasets become more comprehensive and model architectures continue to advance, AI inspection technologies are expected to play an increasingly critical role in reducing production costs, improving garment quality, enhancing manufacturing competitiveness, and accelerating the adoption of intelligent automation throughout the global apparel industry. The convergence of AI inspection, intelligent automation, and advanced manufacturing concepts provides a transformative pathway toward the next generation of garment production systems, reinforcing the industry's transition from labour-intensive operations to data-driven, intelligent, and highly efficient manufacturing environments.